\documentclass[10pt,twoside]{article}
\usepackage[utf8]{inputenc}
\usepackage{amssymb}
\usepackage{float}
\usepackage{algorithm}
\usepackage{algorithmic}
\usepackage{ragged2e}
\usepackage{pifont}
\usepackage{dsfont}
\usepackage[unicode]{hyperref}
\usepackage{amsfonts}
\usepackage[centertags]{amsmath}
\usepackage{amsthm}
\usepackage{newlfont}
\usepackage{array}
\usepackage{graphicx}
\usepackage{calrsfs}
\usepackage{xcolor}
\usepackage[mathscr]{euscript}

\begin{document}

\centerline{\Large{\bf Renewable Energy Sources Selection Analysis}}
\centerline{\Large{\bf  with the Maximizing Deviation Method}}

\centerline{}

\centerline{\bf \bf {Murat Kiri\c{s}ci*}}
\centerline{}

\centerline{ Istanbul University-Cerrahpa\c{s}a, Department of Biostatistics and Medical Informatics, Istanbul, T\"{u}rkiye}
\centerline{e-mail: murat.kirisci@iuc.edu.tr}

\centerline{*Corresponding Author}
\centerline{}

\newtheorem{Theorem}{\quad Theorem}[section]

\newtheorem{Definition}[Theorem]{\quad Definition}

\newtheorem{Proposition}[Theorem]{\quad Proposition}

\newtheorem{Corollary}[Theorem]{\quad Corollary}

\newtheorem{Lemma}[Theorem]{\quad Lemma}

\newtheorem{Example}[Theorem]{\quad Example}

\centerline{}
{\textbf{Abstract:}
Multi-criteria decision-making methods provide decision-makers with appropriate tools to make better decisions in uncertain, complex, and conflicting situations. Fuzzy set theory primarily deals with the uncertainty inherent in human thoughts and perceptions and attempts to quantify this uncertainty. Fuzzy logic and fuzzy set theory are utilized with multi-criteria decision-making methods because they effectively handle uncertainty and fuzziness in decision-makers' judgments, allowing for verbal judgments of the problem. This study utilizes the Fermatean fuzzy environment, a generalization of fuzzy sets. An optimization model based on the deviation maximization method is proposed to determine partially known feature weights. This method is combined with interval-valued Fermatean fuzzy sets. The proposed method was applied to the problem of selecting renewable energy sources. The reason for choosing renewable energy sources is that meeting energy needs from renewable sources, balancing carbon emissions, and mitigating the effects of global climate change are among the most critical issues of the recent period. Even though selecting renewable energy sources is a technical issue, the managerial and political implications of this issue are also important, and are discussed in this study.
\centerline{}



\section{Introduction}

Energy is essential to human life and greatly influences a nation's economic development. However, overuse of fossil fuels causes major environmental issues, including a rise in greenhouse gas emissions that contribute to climate change and global warming. To address environmental problems, most nations are actively developing renewable energy sources ($\mathcal{RES}$). As seen by the 1997 Kyoto Protocol discussions and the completion of the 2015 Paris Climate Change Conference, many countries are dedicated to cutting carbon emissions and creating a green economy because they understand the serious threat that climate change poses. Therefore, many countries must move away from fossil fuels and toward sustainable energy \cite{leeChang}. Energy is becoming more and more necessary for nations to continue developing.\\

However, preserving natural systems while generating the necessary energy is equally crucial. Since the 1990s, nations, international organizations, and institutions have emphasized producing energy in an environmentally sustainable manner. With increased environmental awareness, $\mathcal{RE}$ has gained much more importance. During their natural cycle, $\mathcal{RES}$ can either naturally regenerate themselves over time or continuously renew themselves. Until recently, objective conditions and the uncompetitiveness (high cost) of equipment for large-scale applications hindered the integrated utilization of $\mathcal{RES}$. Nevertheless, the difficulty of boosting the effectiveness of employing the current renewable resources will escalate as solar cell prices steadily decline and tariffs on organic fuels and energy services increase. The policy of a gradual shift to a carbon-neutral economy further aids this.\\

The installed capacity of RE is 15.1 GW, and the total investment in 2024 will exceed 728 billion US dollars globally, according to the REN21 Global Status Report 2025. The average yearly growth rate is 0.88\% \cite{GSR2025}. However, renewable sources accounted for the largest share of the increase in the total energy supply (38\%). Only a 1\% rise in the energy intensity of the global economy prolonged the recent downturn. The growth in energy-related CO2 emissions was 0.8\% lower than the 1.2\% increase in 2023 \cite{GER2025}. According to IRENA, $\mathcal{RE}$ will account for roughly 40\% of global energy by 2030 due to decreased technology prices. $\mathcal{RE}$ is infinite, pure energy with many benefits over conventional fossil fuel energy, such as plentiful and cost-free. However, because solar rays and wind power are unpredictable, $\mathcal{RE}$ has limitations regarding output and capacity. Specifically, the price of electricity produced using $\mathcal{RES}$s is now higher than that produced using fossil fuels. \\

The $\mathcal{RE}$ Policy Network for the 21st Century website \cite{GSR2025} provides various statistics on this topic. Some of these statistical graphs are shown in Figures \ref{fig10} and \ref{fig20}.

\begin{figure}[htb!]
	\centering
	\includegraphics[width=15cm,height=6cm]{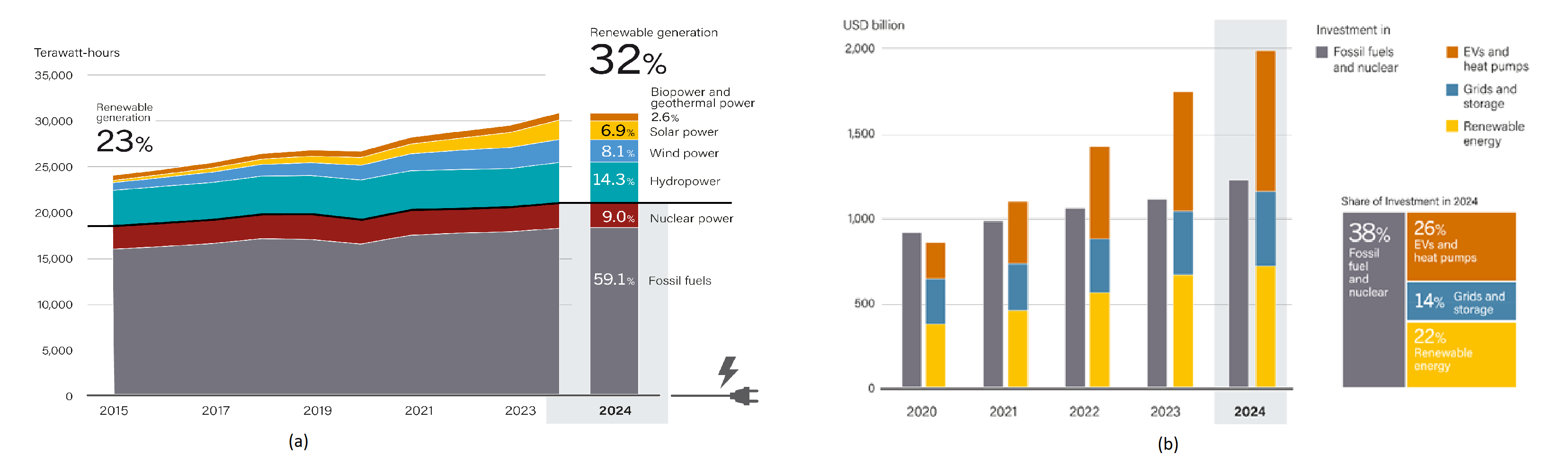}
	\caption{(a) Electricity Generation by Energy Source, 2015-2024;  (b) Global Investment in Selected Energy Technologies, 2020-2024}\label{fig10}
\end{figure}

\begin{figure}[htb!]
	\centering
	\includegraphics[width=16cm,height=5cm]{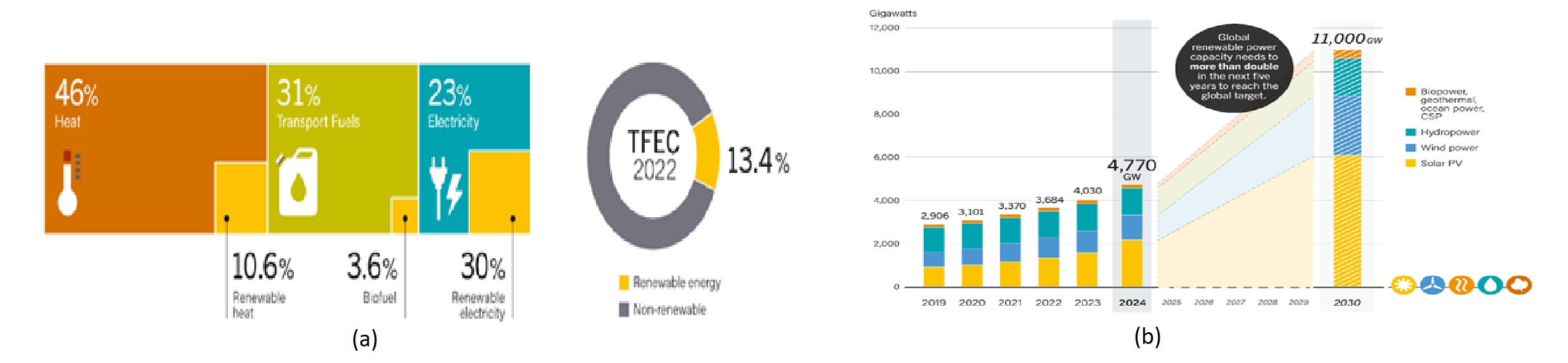}
	\caption{(a) Total Final Energy Consumption (TFEC) and Share of Modern Renewable Energy, by Energy Carrier, 2022;  (b) Renewable Power Capacity by Technology, 2019-2024, Compared to 2030 Global Tripling Target}\label{fig20}
\end{figure}

The selection of $\mathcal{RES}$s is a multi-criteria challenge with several conflicting factors. When deciding on the selection criteria, we must consider the superiority and limitations of each option. To solve this problem, a multi-criteria strategy should be employed. Such criteria provide the evaluation process with more uncertainty and complexity. It can be difficult for decision-makers ($\mathcal{DMR}$) to give whole numerical values for the criteria or attributes, and their evaluations are frequently imprecise in many decision-making($\mathcal{DM}$) scenarios. $\mathcal{DMR}$s usually use language to convey the assessment data of the alternative energy policy's acceptability for many subjective criteria and the weights, because most evaluation factors are difficult to quantify precisely. Fuzzy logic is an effective way to imitate this ambiguity in human preferences. Fuzzy logic is a branch of fuzzy set theory that focuses on handling imprecise input by utilizing membership functions. In 1965, Zadeh formalized fuzzy set theory \cite{Zadeh}.\\

Fuzzy assessment information is fused utilizing aggregation operators to help organize all the options in real-world multi-criteria $\mathcal{DM}$($\mathcal{MCDM}$) situations. Assigning the attribute's weight is a crucial issue since we cannot always collect comprehensive information in the complicated $\mathcal{DM}$ environment, and the attribute's weight is always partially or entirely unknown. In several $\mathcal{DM}$ domains, similar multi-criteria $\mathcal{DM}$ ($\mathcal{MCDM}$) problems have been addressed using Wang's  \cite{wang} maximizing deviation ($\mathcal{MD}$) technique. As we all know, ranking the fused values makes it simple to choose the superior options. When a characteristic significantly affects ranking outcomes, it is evident that this feature is crucial in affecting decision results and should be given significant weight to illustrate the differences between all the options clearly. Conversely, if an attribute minimizes how options are arranged, it is irrelevant to the $\mathcal{DM}$ process and should therefore be given less weight. Put another way, we should give the option that makes a wider deviation more weight and the option that makes a lesser difference less weight.\\

When many uncertainties and judgments can or should be made sequentially, robustness analysis offers a method for organizing the problem. The robustness analysis focuses on preserving flexibility by using the difference between plans and decisions. Generally speaking, robustness tells us how flexible the options are.  \\

During method development or validation, the robustness check can be performed at any point. Robustness is both a legal requirement and a sign of best practices in method development. Method validation can be supported by robustness data gathered during development. This study will perform robustness analysis using the COPRAS method based on IVFF. The  Complex Proportional Assessment (COPRAS) method is one of the $\mathcal{MCDM}$ methods and was developed to help $\mathcal{DMR}$s select the most suitable alternative from a set of alternatives. This methodology is often used to rank and evaluate alternatives that must be assessed based on various criteria. Specifically, COPRAS allows $\mathcal{DMR}$s to evaluate each alternative's relative benefits (or harms) by comparing them against each other. The fundamental superiority of COPRAS is that it can determine both benefit and harm criteria.

\subsection{Necessity}

By lowering carbon dioxide emissions, $\mathcal{RES}$s contribute to environmental protection, one of their most important features. Because they are domestically sourced, they contribute to reducing external energy dependency and increasing employment. They also attract widespread and strong public support. In other words, $\mathcal{RES}$s possess all the characteristics of accessibility, availability, and acceptability.\\

The importance of $\mathcal{RES}$s is increasing today. These sources represent energy derived from natural resources that have less environmental impact and are infinitely renewable. Unlike nuclear and fossil fuels, $\mathcal{RES}$s contribute significantly to the fight against climate change. Renewable energy sources, including geothermal, hydro, wind, and solar, reduce carbon emissions without endangering natural ecosystems. \\

Furthermore, $\mathcal{RES}$s are critical to a sustainable energy future. Because they are derived from inexhaustible natural resources, they provide long-term energy security and reduce dependence on energy imports by increasing energy independence. From an economic perspective, the $\mathcal{RE}$ sector is growing. $\mathcal{RE}$ projects create new jobs, contribute to a green economy, and support economic development.\\

In conclusion, $\mathcal{RES}$s represent the key to the global energy transition with their environmental, economic, and social benefits. They play a critical role in ensuring a sustainable life for future generations.\\

There is a significant amount of uncertainty because many $\mathcal{RES}$s are still in their infancy in developing nations. Effective choices must be made, and a suitable procedure for choosing the best sources is required. A crucial step in the $\mathcal{DM}$ process, particularly when developing decision models, is assessing the current criteria and choosing the best available resources. This research addresses the primary issue: "Once the criteria to be considered in the selection of $\mathcal{RES}$ are determined, is it possible to propose a new $\mathcal{DM}$ model that will help $\mathcal{DMR}$s estimate their uncertainties to address the selection of $\mathcal{RES}$s?"


\subsection{Research Gap}
Choosing a suitable technology is essential in selecting $\mathcal{RES}$s, which can be considered a classic $\mathcal{MCDM}$ problem.\\

$\mathcal{RES}$s take their place in our lives as alternative energies because they are resources that exist spontaneously and do not run out over time. In this situation, selecting the optimal management technology is complex and unresolved. Different sources and technologies can be selected in several ways. It is crucial to evaluate which of the options is the best. Fundamental factors, such as environmental, technical,  economic, and social,  should all be considered while evaluating each option. Choosing the most appropriate solution for such situations is a challenging $\mathcal{MCDM}$ problem with various qualitative and quantitative aspects. The suggested methodology tackles the selection of $\mathcal{RES}$s. In order to rate candidates, model the connected vagueness, and optimize their benefits, this study uses a $\mathcal{MD}$ method based on IVFFSs.\\

When applied to the selection of $\mathcal{RES}$s, the $\mathcal{MD}$ method has the potential to fill some gaps in the literature:\\

- It will address the lack of diversity in $\mathcal{DM}$ methods. The literature extensively uses techniques like TOPSIS, AHP, VIKOR, WASPAS, and COPRAS. However, since the $\mathcal{MD}$ method is less preferred, a study using this method will demonstrate the applicability of an alternative $\mathcal{MCDM}$ technique to the energy selection problem.\\

- By reflecting uncertainty and differences between criteria, the method maximizes the differences between criteria, bringing a unique perspective to the $\mathcal{DM}$ process. This allows for clearer identification of distinguishing differences between alternatives that might otherwise be overlooked in traditional methods.\\

- $\mathcal{RE}$ selection, in terms of group $\mathcal{DM}$ and $\mathcal{DMR}$ opinions, is generally a multi-actor issue (government, investors, environmentalists, academia). A $\mathcal{MD}$ approach can more objectively reflect the differences in criteria in group $\mathcal{DM}$ settings. This could address the lack of a more balanced integration of $\mathcal{DMR}$ opinions in the literature.\\

- Given regional or country-specific implementation gaps, standardized methods are often used in studies on developing countries like T\"{u}rkiye or regional energy policies. An analysis using $\mathcal{MD}$ can reveal how country-specific criteria are weighted differently, providing a fresh perspective for policy decision support systems.\\

- A more balanced approach to sustainability by simultaneously balancing environmental, economic, social, and technical criteria may yield different results. This closes a gap in the literature on analyses that rely on classical scoring or unidimensional weightings.

\subsection{Contribution}

The study's noteworthy contributions are highlighted by the following points:\\

- Proposing a new $\mathcal{MD}$ method based on IVFFSs to solve the $\mathcal{MCDM}$ problems.\\
-Given the algorithm of the new method.\\
- Analysed $\mathcal{MCDM}$ problem of selecting $\mathcal{RES}$s with IVFF-$\mathcal{MD}$ method.\\
- Making robustness analysis. \\
- The suggested methodology will be verified and validated through comparative evaluations using various techniques.

\section{Literature Review:}

Zadeh's \cite{Zadeh} fuzzy set (FS) notion is essential in the mathematical technique of fuzzy modeling, which describes uncertainty in human systems. However, this technique is insufficient to fully express human judgments. In response to this weakness, Atanassov \cite{Atan} created the intuitionistic fuzzy set (IFS) theory. IFS is not designed to manage situations where the sum of MD and ND for some alternatives exceeds one(\cite{boranRES}, \cite{dumrul}, \cite{kirisciComp} \cite{weiMDM}). Yager \cite{Yager0}, \cite{Yager} developed Pythagorean fuzzy sets (PFSs) to get around this restriction by loosening it up so that the only condition at any evaluation of an option is that the total of the squares of MD and ND is less than 1(\cite{hosse}, \cite{kirisciSIGMA}, \cite{kirisciVIKOR}, \cite{Liangetal}). Senapati and Yager (\cite{SenYager}, \cite{SenYager1}) developed the FFS theory in response to the limitations imposed by PFSs. In response to the limits imposed by IFSs and PFSs, Senapati and Yager (\cite{SenYager}, \cite{SenYager1}) introduced the FFS theory. In an FFS, the cubic total of membership and non-membership degrees must be less than or equal to one. In addition, FFS-related applications are depicted in \cite{alkan, gargetal, Jeevaraj}, \cite{kirisci0}- \cite{LiuLiu}, \cite{SenYager1, SenYager2, kirisci22}.\\

Numerous studies have ranked RES using $\mathcal{MCDM}$ methodologies. These methods are instrumental when selecting the ideal RES, necessitating balancing several complex considerations. Kolagar et al. proposed a hybrid strategy that combines DEA and the fuzzy best-worst method to prioritize RES in Iran \cite{kol}.  An integrated system comprising fuzzy-TOPSIS, AHP, and Slack-based DEA was created by Xu et al. \cite{XuRES} in order to rank and prioritize Pakistan's most sustainable and effective hydrogen generation techniques. The economic, technological, political, social, and environmental aspects of sustainability are all included in this approach. Jha et al. \cite{jha} focused on utilizing an Energy Index metric and a fuzzy AHP to rank $\mathcal{RE}$ options in India. This index was created by combining the scores of several solutions based on eleven environmental and techno-economic factors. Solangi et al. \cite{solangi} evaluated several potential sites for solar power installations using fuzzy VIKOR and AHP algorithms. Fuzzy VIKOR was used to sort viable solutions after each criterion and sub-criterion was weighed using AHP. Shah et al. \cite{shah} used a combination of fuzzy Delphi, fuzzy AHP, and environmental DEA approaches to investigate the viability of various RES for hydrogen production in Pakistan. Geothermal, biomass, wind, solar, micro-hydro, and municipal solid waste  were the six $\mathcal{RES}$ choices that the researchers concentrated on. In a study by Longsheng et al. \cite{long}, the SWOT factors and strategies were completed by literature research. The twelve SWOT-derived strategies were then ranked using Grey-TOPSIS, utilizing the FAHP weights to offer a systematic technique to strategy assessment. The existing literature shows that the FAHP, FDEA techniques effectively select and rank RES. Kahraman et al. \cite{kahramanRES} introduced an $\mathcal{MCDM}$ methodology based on axiomatic design and AHP techniques to assist nations in selecting which $\mathcal{RES}$ to invest in. In \cite{boranRES}, $\mathcal{RE}$ technologies for power production in Turkey were evaluated using intuitionistic fuzzy TOPSIS. Lee and Chang presented a new $\mathcal{MCDM}$ method combining WSM, VIKOR, TOPSIS, and ELECTRE \cite{leeChang}. They utilized Shannon's entropy weighted technique to rank $\mathcal{RES}$s, assessing the importance of each aspect. Using a hybrid $\mathcal{DM}$ methodology that included the TOPSIS-I and BWM techniques and several criteria, the most significant alternatives for sustainable $\mathcal{RE}$ were identified in \cite{als}. The model proposed by Öztürk et al. \cite{ozturk} to assist $\mathcal{DMR}$s in selecting $\mathcal{RES}$s comprises AHP and VIKOR. This model shows how the criteria affect Chicago's $\mathcal{RE}$ options' performance metrics and priority. The best $\mathcal{RE}$ options in Turkey are evaluated in \cite{dumrul} using the intuitionistic fuzzy EDAS technique. Husain et al. \cite{husain} presented a new and trustworthy integrated approach that combines the entropy, CRITIC, PIV, MABAC, MARCOS, WASPAS, MOORA, EDAS, GRA, and TOPSIS methodologies to aid in $\mathcal{DM}$ in the field of sustainable energy. In order to evaluate $\mathcal{RES}$s in impoverished countries, Dehshiri et al. \cite{hosse} suggested a decision framework that included BWM and Interval-Valued Pythagorean Fuzzy WASDPAS approaches. Luhaniwal et al. \cite{luh} investigate various $\mathcal{RES}$s in India, including solar, geothermal, hydro, biomass, wave, onshore, and offshore wind energy, using an integrated DEA and fuzzy AHP approach. Shamsipour et al. \cite{sham} evaluated five $\mathcal{RE}$ technologies in Isfahan: solar photovoltaic, biomass, wind, concentrated solar power, and geothermal energy. They did this by combining fuzzy DEMATEL-ANP and fuzzy MULTIMOORA algorithms. Wang et al. \cite{wangetal} established an $\mathcal{MCDM}$ to examine energy policies and investments in $\mathcal{RES}$s. First, a balanced scorecard technique examines the criteria using a correlation coefficient according to interval type-2 FSs. Then, using interval type-2 FSts, a FTOPSIS and FDEMATEL approach is used.\\


Wu and Chen \cite{wuchen} investigated $\mathcal{MCDM}$ problems with linguistic information using the $\mathcal{MD}$ method. Wei \cite{weiMDM} examined the $\mathcal{MD}$ approach in an intuitionistic fuzzy setting. Liang, Zhang, and Liu \cite{Liangetal} employed the $\mathcal{MD}$ approach to address $\mathcal{MCDM}$ problems and proposed aggregation operators under the interval-valued Pythagorean fuzzy environment. Using the $\mathcal{MD}$ approach, Sahin and Liu \cite{sahin} developed two novel neutrosophic models and used them to solve $\mathcal{MCDM}$ problems. Selvachandran, Quek, Smarandache, and Broumi \cite{selva}, who used the $\mathcal{MD}$ approach for $\mathcal{MCDM}$, proposed the single-valued neutrosophic TOPSIS model. By utilizing neutrosophic information to $\mathcal{MD}$ approach, Xiong and Cheng \cite{xiong} proposed a novel method for calculating the weights of the qualities. Pamucar, Sremac, Cirovic, and Tomic \cite{pametal}, who also used the $\mathcal{MD}$ approach to establish the attribute weights, constructed the LNN WASPAS model.


\section{Preliminaries}\label{chap:1}


\begin{Definition}\cite{Jeevaraj}
Let $SUB[0,1]$ be a closed subinterval of the unit interval. An IVFFS on $E \neq \emptyset$ is an expression given by $F=\{(k, [\zeta_{F_{L}}(k), \zeta_{F_{U}}(k)], [\eta_{F_{L}}(k), \eta_{F_{U}}(k)]): k \in E\}$, where $\zeta_{F}(k), \eta_{F}(k) \in SUB[0,1]$ with $0 < \sup_{k}(\zeta_{F}(k))^{3}+ \sup_{k}(\eta_{F}(k))^{3} \leq 1$.
\end{Definition}

The equation $\theta_{F}=[\theta_{FL}, \theta_{FU}]=[(1-\zeta_{FU}^{3}-\eta_{FU}^{3})^{1/3}, (1-\zeta_{FL}^{3}-\eta_{FL}^{3})^{1/3}]$ is said to be a hesitation degree.

\begin{Definition}\cite{Jeevaraj}\label{def102}
	For IVFFSs $F=([\zeta_{F_{L}}(a), \zeta_{F_{U}}(a)], [\eta_{F_{L}}(a), \eta_{F_{U}}(a)])$,  $F_{1}=([\zeta_{F_{1L}}(a), \zeta_{F_{1U}}(a)], [\eta_{F_{1L}}(a), \eta_{F_{1U}}(a)])$, $F_{2}=([\zeta_{F_{2L}}(a), \zeta_{F_{2U}}(a)], [\eta_{F_{2L}}(a), \eta_{F_{2U}}(a)])$,
	
	\begin{itemize}
		\item $F_{1} \cup F_{2} = \left( [\max(\zeta_{F_{1L}}, \zeta_{F_{2L}}), \max(\zeta_{F_{1U}}, \zeta_{F_{2U}})],   [\min(\eta_{F_{1L}}, \eta_{F_{2L}}), \min(\eta_{F_{1U}}, \eta_{F_{2U}})]\right)$
		\item $F_{1} \cap F_{2} = \left( [\min(\zeta_{F_{1L}}, \zeta_{F_{2L}}), \min(\zeta_{F_{1U}}, \zeta_{F_{2U}})],   [\max(\eta_{F_{1L}}, \eta_{F_{2L}}), \max(\eta_{F_{1U}}, \eta_{F_{2U}})]\right)$
		\item $F^{c} = ([\eta_{F_{L}}, \eta_{F_{U}}], [\zeta_{F_{L}}, \zeta_{F_{U}}])$
		\item $F_{1} \oplus F_{2} = \Bigg(  \Big[\sqrt[3]{(\zeta_{F_{1L}})^{3}+(\zeta_{F_{2L}})^{3}-(\zeta_{F_{1L}})^{3}.(\zeta_{F_{2L}})^{3} }, \\
		\sqrt[3]{(\zeta_{F_{1U}})^{3}+(\zeta_{F_{2U}})^{3}-(\zeta_{F_{1U}})^{3}.(\zeta_{F_{2U}})^{3} }  \Big],    \left[\eta_{F_{1L}}\eta_{F_{2L}}, \eta_{F_{1U}}\eta_{F_{2U}}  \right]   \Bigg)$
		\item $F_{1} \otimes F_{2} = \Bigg(  \left[\zeta_{F_{1L}}\zeta_{F_{2L}}, \zeta_{F_{1U}}\zeta_{F_{2U}}  \right],   \Big[\sqrt[3]{(\eta_{F_{1L}})^{3}+(\eta_{F_{2L}})^{3}-(\eta_{F_{1L}})^{3}.(\eta_{F_{2L}})^{3} },\\
		\sqrt[3]{(\eta_{F_{1U}})^{3}+(\eta_{F_{2U}})^{3}-(\eta_{F_{1U}})^{3}.(\eta_{F_{2U}})^{3} }  \Big]    \Bigg)$
		\item $\lambda F = \left(  \left[\sqrt[3]{1- \left(1- \zeta_{F_{L}}^{3}\right)^{\lambda}}, \sqrt[3]{1- \left(1- \zeta_{F_{U}}^{3}\right)^{\lambda}}\right],  \left[\eta_{F_{L}}^{\lambda}, \eta_{F_{U}}^{\lambda}\right]  \right)$
		\item $F^{\lambda}= \left( \left[\zeta_{F_{L}}^{\lambda}, \zeta_{F_{U}}^{\lambda}\right],  \left[\sqrt[3]{1- \left(1- \eta_{F_{L}}^{3}\right)^{\lambda}}, \sqrt[3]{1- \left(1- \eta_{F_{U}}^{3}\right)^{\lambda}}\right] \right)$
	\end{itemize}
\end{Definition}

\begin{Definition}\cite{ranMis1}
	For the IVFFS $F=([\zeta_{FL}(k), \zeta_{FU}(k)], [\eta_{FL}(k), \eta_{FU}(k)])$, the equations \ref{scr}, \ref{acc}, and \ref{Nscr} is said to be functions of score, accuracy and normalized score.
	\begin{eqnarray}\label{scr}
		\mathcal{SC}(F)&=&1/2\left( [(\zeta_{FL}(k))^{3}+(\zeta_{FU}(k))^{3}]-[(\eta_{FL}(k))^{3}+(\eta_{FU}(k))^{3}] \right) \\ \label{acc}
		\mathcal{AC}(F)&=&1/2\left( [(\zeta_{FL}(k))^{3}+(\zeta_{FU}(k))^{3}]+[(\eta_{FL}(k))^{3}+(\eta_{FU}(k))^{3}] \right) \\ \label{Nscr}
		\overline{\mathcal{SC}}(F)&=&\frac{\left(\mathcal{\mathcal{SC}}(F)+1 \right)}{2}
	\end{eqnarray}
\end{Definition}

It is note that $ \mathcal{SC}(F) \in [-1, 1] $,  $ \mathcal{AC}(F) \in [0, 1] $, and $ \overline{\mathcal{SC}}(F) \in [0, 1] $.

IVFF weighted averaging(IVFFWA) and geometric(IVFFWG) operators are characterized by

\begin{eqnarray}\label{IVFFWA}
		&& IVFFWA( F_{1}, F_{2}, \cdots, F_{n}) \\ \nonumber
	&=& \Bigg(\Bigg[ \sqrt[3]{\left(1-\prod_{i=1}^{{n}}\left(1-\zeta_{F_{iL}}^{3}\right)^{\omega_{i}}\right)}, \sqrt[3]{\left(1-\prod_{i=1}^{{n}}\left(1-\zeta_{F_{iU}}^{3}\right)^{\omega_{i}}\right)}\Bigg], 	\Bigg[\prod_{i=1}^{{n}}\left(\eta_{F_{iL}}\right)^{\omega_{i}},  \prod_{i=1}^{{n}}\left(\eta_{F_{iU}}\right)^{\omega_{i}}\Bigg]	  \Bigg), \\ \label{IVFFWG}
	&& IVFFWG( F_{1}, F_{2}, \cdots, F_{n}) \\ \nonumber
	&=& \Bigg( \Bigg[\prod_{i=1}^{{n}}\left(\zeta_{F_{iL}}\right)^{\omega_{i}},  \prod_{i=1}^{{n}}\left(\zeta_{F_{iU}}\right)^{\omega_{i}}\Bigg],
	\Bigg[ \sqrt[3]{\left(1-\prod_{i=1}^{{n}}\left(1-\eta_{F_{iL}}^{3}\right)^{\omega_{i}}\right)}, \sqrt[3]{\left(1-\prod_{i=1}^{{n}}\left(1-\eta_{F_{iU}}^{3}\right)^{\omega_{i}}\right)}\Bigg]  \Bigg).
\end{eqnarray}
where $\omega_{k}$ is given as the influence weight.


\section{Method}
\subsection{IVFF-$\mathcal{MD}$ Method }

The weights of the criteria are decided by the $\mathcal{MD}$ model. The decision can then be taken once the IVFFWA operator is shown to aggregate the provided decision information. Lastly, the suggested method's algorithm is presented.\\

Firstly, let's set up the $\mathcal{MCDM}$ problem in the IVFF environment:\\

A discrete set of alternatives: $S=\{S_{1}, S_{2}, \cdots, S_{m}\}$.\\
A finite set of criteria: $K=\{K_{1}, K_{2}, \cdots, K_{n}\}$.\\
A set of $\mathcal{DMR}$s: $U=\{U_{1}, U_{2}, \cdots, U_{g}\}$.\\
Let $\lambda=\{\lambda_{1}, \lambda_{2}, \cdots, \lambda_{g}\}$ show the influence weights of $\mathcal{DMR}$s ($0 \leq \lambda_{k} \leq 1$ and $\sum_{k=1}^{g} \lambda k=1$).\\
Let $U^{k}$: $\omega^{k} = (\omega_{1}^{k}, \omega_{2}^{k}, \cdots, \omega_{n}^{k} )^{T}$ indicate the weight vectors of criteria for the $\mathcal{DMR}$.\\


The $k^{th}$ $\mathcal{DMR}$ expresses the $j^{th}$ criterion of the $i^{th}$ alternative in the problem with IVFFN $F_{ij}^{k}$.


\begin{Definition}
The IVFF decision matrix($\mathcal{DCSM}$) is defined by $P^{k}=(F_{ij}^{k})_{m \times n}=([\zeta^{L}_{(F_{ij}^{k})}, \zeta^{U}_{(F_{ij}^{k})}], [\eta^{L}_{(F_{ij}^{k})}, \eta^{U}_{(F_{ij}^{k})}])_{m \times n}$, $k \in \{1,2,\cdots,g$ if all entries of the matrix $P^{k}$ are IVFFNs.
\end{Definition}


Now, let us set up the $\mathcal{MD}$ model to determine the optimal weights.\\

The $\mathcal{MD}$ method, originally proposed by Wang \cite{wang}, is used to determine the weights of criteria for solving $\mathcal{MCDM}$ problems with crisp (non-fuzzy) numbers. In this research, an optimization model for identifying the ideal criteria weights in an IVFF context is established using the main structure of the $\mathcal{MD}$ approach. First, we compute the deviations between each alternative and the others using the IVFF-distance metric.

\begin{Definition}
	Equation \ref{deviation1} is called the deviation value between the alternative $S_{\xi}$ and the alternative $S_{\sigma}$ $(\xi \neq \sigma)$:
\begin{eqnarray}\label{deviation1}
  &&D_{\xi \sigma j}^{k} = \omega_{j}^{k}DIST\left(F_{\xi j}^{k}, F_{\sigma j}^{k} \right)=\frac{1}{4}\omega_{j}^{k} \Bigg( \left| (\hat{m}_{F_{\xi j}}^{-}(k))^{3}-(\hat{m}_{F_{\sigma j}}^{-}(k))^{3}\right| \\ \nonumber
  &&+\left| (\hat{m}_{F_{\xi j}}^{+}(k))^{3}-(\hat{m}_{F_{\sigma j}}^{+}(k))^{3}\right|
   + \left| (\hat{n}_{F_{\xi j}}^{-}(k))^{3}-(\hat{n}_{F_{\sigma j}}^{-}(k))^{3}\right|  +\left| (\hat{n}_{F_{\xi j}}^{+}(k))^{3}-(\hat{n}_{F_{\sigma j}}^{+}(k))^{3}\right|\\ \nonumber
  && + \left| (\theta_{F_{\xi j}}^{-}(k))^{3}-(\theta_{F_{\sigma j}}^{-}(k))^{3}\right| +\left| (\theta_{F_{\xi j}}^{+}(k))^{3}-(\theta_{F_{\sigma j}}^{+}(k))^{3}\right|   \Bigg),
\end{eqnarray}
where $\mathcal{DMR}$ $U_{k}$ $(k=1,2,\cdots,g)$ and  the criterion $K_{j}$ $(j=1,2,\cdots,n)$.
\end{Definition}

Then, for the criterion $K_{j}$ $(j=1,2,\cdots,n)$ and the $\mathcal{DMR}$ $U_{k}$ $(k=1,2,\cdots,g)$, the deviation value between the alternative $S_{i}$ $(i=1,2,\cdots,m)$ and all the other alternatives can be calculated as:

\begin{eqnarray}
  D_{\xi j}^{k} = &=& \sum_{\sigma=1}^{m}\omega_{j}^{k}DIST\left(F_{\xi j}^{k}, F_{\sigma j}^{k} \right).
\end{eqnarray}

Furthermore, the deviation value of all the alternatives to the other alternatives can be calculated as:

\begin{eqnarray}
  D_{j}^{k} = &=& \sum_{\xi=1}^{m}\sum_{\sigma=1}^{m}\omega_{j}^{k}DIST\left(F_{\xi j}^{k}, F_{\sigma j}^{k} \right).
\end{eqnarray}

A criterion has a lesser significance in the prioritizing process for an $\mathcal{MCDM}$ problem if the criterion values of all alternatives under a criterion differ just a little. On the other hand, this criterion is more crucial in selecting the best option if there are discernible disparities in all choices' criterion values. In other words, while ranking the alternatives, a criterion should be given a lower weight if its values are similar across alternatives; if not, a larger weight should be given to the criterion that deviates from the norm, regardless of its importance. Specifically, most $\mathcal{DMR}$s will consider a criterion insignificant and give it zero weight if all alternatives receive the same score on that criterion. \\

In order to choose the weight vector $\ omega^{k}$ for the $\mathcal{DMR}$ $U_{k}$ $(k=1,2,\cdots,g)$, we create an ideal model that optimizes all deviation values for every criterion as follows:

\begin{eqnarray} \label{st}
\left\{\begin{array}{ll}
&\max D^{k} =  \sum_{j=1}^{n}\sum_{\xi=1}^{m}\sum_{\sigma=1}^{m}\omega_{j}^{k}DIST\left(F_{\xi j}^{k}, F_{\sigma j}^{k} \right)\\
s.t. & \sum_{j=1}^{n}\left(\omega_{j}^{k}\right)^{3}=1,\\
& \omega_{j}^{k} \geq 0, \quad j=1,2,\cdots,n.
	\end{array} \right.
\end{eqnarray}
 The Lagrange function of the optimization Model \ref{st} can be obtained as:

\begin{eqnarray}
\mathfrak{L}(\omega^{k}, \lambda) = \sum_{j=1}^{n}\sum_{\xi=1}^{m}\sum_{\sigma=1}^{m} \omega_{j}^{k}DIST\left(F_{\xi j}^{k}, F_{\sigma j}^{k} \right) + \frac{\lambda}{3} \left( \sum_{k=1}^{n}(\omega_{j}^{k}) ^{3} - 1 \right)
\end{eqnarray}

where The Lagrange multiplier variable is represented by the real number $\lambda$.\\

If we take the partial derivative for $\mathfrak{L}$,

\begin{eqnarray}\label{deriv}
\frac{\partial \mathfrak{L}}{\partial \omega_{j}^{k}}&=&\sum_{\xi=1}^{m}\sum_{\sigma=1}^{m}DIST\left(F_{\xi j}^{k}, F_{\sigma j}^{k} \right)+\lambda.(\omega_{j}^{k})^{2}=0\\ \nonumber
\frac{\partial \mathfrak{L}}{\partial \lambda}&=& \frac{1}{3}\left( \sum_{j=1}^{n}(\omega_{j}^{k}) ^{3} - 1 \right)=0.
\end{eqnarray}

Using \ref{deriv},
\begin{eqnarray}
 \omega_{j}^{k} = \frac{\sum_{\xi=1}^{m}\sum_{\sigma=1}^{m}DIST\left(F_{\xi j}^{k}, F_{\sigma j}^{k} \right)}{\sqrt{\sum_{j=1}^{n}\left(\sum_{\xi=1}^{m}\sum_{\sigma=1}^{m}DIST\left(F_{\xi j}^{k}, F_{\sigma j}^{k} \right)\right)^{2}}}.
\end{eqnarray}

Finally, the optimal weight $\omega_{j}^{k}$ is normalized as:

\begin{eqnarray}\label{optimal}
\overline{ \omega}_{j}^{k} = \frac{\omega_{j}^{k}}{\sum_{j=1}^{n} \omega_{j}^{k}}.
\end{eqnarray}

We can create a different programming model if the attribute weight information is entirely unknown:
\begin{eqnarray}\label{model2}
\left\{\begin{array}{ll}
&\max D^{k} =  \sum_{j=1}^{n}\sum_{\xi=1}^{m}\sum_{\sigma=1}^{m}\omega_{j}^{k}DIST\left(F_{\xi j}^{k}, F_{\sigma j}^{k} \right)\\
s.t. &  \\
& \sum_{j=1}^{n}\omega_{j}^{k}=1, \quad \omega_{j}^{k} \geq 0, \quad j=1,2,\cdots,n.
	\end{array} \right.
\end{eqnarray}

Model \ref{model2} is solved to find the optimal solution for $\omega^{k}=(\omega^{k}_{1}, \omega^{k}_{2}, \cdots, \omega^{k}_{n})^{T}$.\\

We must ascertain the weights of the criteria for the group after acquiring weights of the criteria for each $\mathcal{DMR}$. Indicate the group's weight for the criterion $K_{j}$ using $\omega_{j}^{*}$. Then, to identify the ideal weights of criteria for the group, we further develop an optimal model that minimizes consistency as follows:

\begin{eqnarray}\label{model3}
\left\{\begin{array}{ll}
&\min E(\omega^{*}) =  \sum_{j=1}^{n}\sum_{k=1}^{g} \alpha_{k}|\omega_{j}^{k} - \omega_{j}^{*}|\\
s.t. &  \\
& \sum_{j=1}^{n}\omega_{j}^{*}=1, \quad \omega_{j}^{*} \geq 0, \quad j=1,2,\cdots,n.
	\end{array} \right.
\end{eqnarray}

Model \ref{model3} solve and
\begin{eqnarray}\label{solve}
\phi_{j}^{k}=\frac{1}{2}\left(|\omega_{j}^{k} - \omega_{j}^{*}|+(\omega_{j}^{k} - \omega_{j}^{*}) \right),\\ \nonumber
\varphi_{j}^{k}=\frac{1}{2}\left(|\omega_{j}^{k} - \omega_{j}^{*}|-(\omega_{j}^{k} - \omega_{j}^{*}) \right).
\end{eqnarray}

The following line programming model is then created from the ideal model \ref{model3}:

\begin{eqnarray}\label{model4}
\left\{\begin{array}{ll}
&\min E(\omega^{*}) =  \sum_{j=1}^{n}\sum_{k=1}^{g} \alpha_{k}(\phi_{j}^{k} + \varphi_{j}^{k})\\
 &  \omega_{j}^{k} - \omega_{j}^{*} - \phi_{j}^{k} + \varphi_{j}^{k}=0,\\
s.t.& \phi_{j}^{k} \geq 0,\\
& \varphi_{j}^{k} \geq 0,\\
& \phi_{j}^{k}\varphi_{j}^{k}=0,\\
& \sum_{j=1}^{n}\omega_{j}^{*}=1, \quad \omega_{j}^{*} \geq 0, \quad j=1,2,\cdots,n.
	\end{array} \right.
\end{eqnarray}

Model \ref{model4} is solved to find the optimal solution for $\omega^{k}=(\omega^{k}_{1}, \omega^{k}_{2}, \cdots, \omega^{k}_{n})^{T}$.\\

At this stage, the alternatives will be ranked.\\

Once the group's ideal weights for the criteria have been determined, we often need to combine the available data to identify the overall preference value of each choice before making a choice. The order of the alternatives will be established using the IFWA (Equation \ref{IVFFWA}) (or IFWG (Equation \ref{IVFFWG})) operator.

\subsection{Algorithm}
The algorithm and flow chart (Figure \ref{new_method}) of the $\mathcal{MD}$ model based on IVFF are:\\

Step 1: create the IVFF-$\mathcal{DCSM}$.\\

Step 2: Solve Equation \ref{optimal} or model \ref{model2} to find the weights of the criterion for each $\mathcal{DMR}$.\\

Step 3: Solve model \ref{model4} to determine the weights of the group's criterion.\\

Step 4: Determine the alternative's overall preference values. $U_{i}$.\\

Step 5: Compare the magnitude of each alternative's comprehensive preference values to get the best ranking order for the alternatives. Then, choose the best alternative.

\begin{figure}[htb!]
  \centering
  \includegraphics[width=10cm,height=8cm]{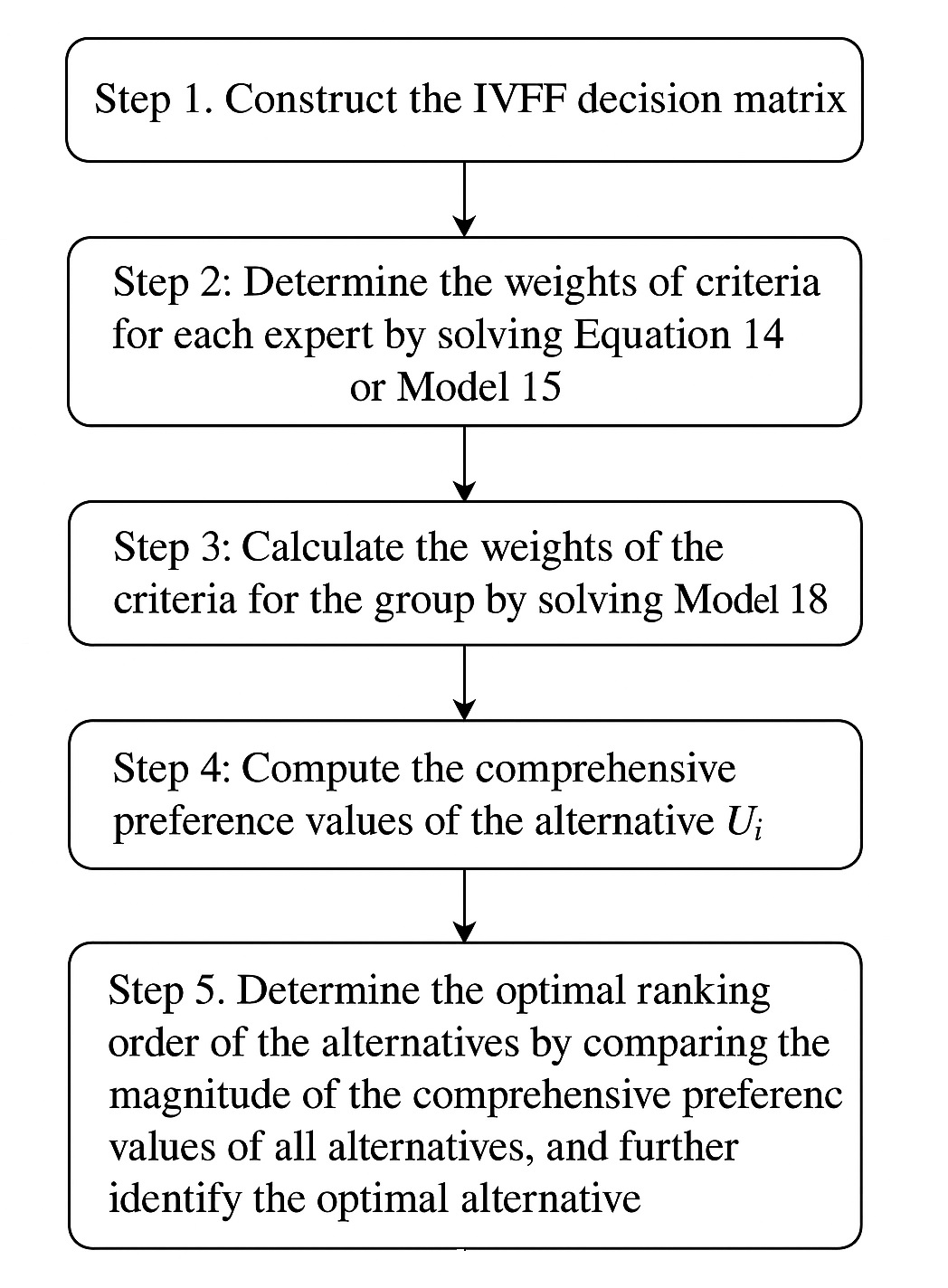}
 \caption{Flow Chart of New Method}\label{new_method}
\end{figure}


\section{An application for Sustainable finance}

	\subsection{Problem Design}
	In addition to the efficient production and consumption of conventional energy sources and the reduction of their environmental impact, the idea of sustainable development—which has grown to be a topic of great concern for all societies worldwide—has highlighted the need to switch to clean and limitless energy sources. $\mathcal{RE}$ research has proliferated due to environmental degradation, global warming, and the growing energy demand primarily satisfied by fossil fuels. It is also predicted that fossil fuels will soon run out. $\mathcal{RE}$ is derived from the flow of energy that occurs naturally while processes continue. $\mathcal{RES}$s are resources like solar, water, wind, and waves that are exploited to generate other energy sources before they run out. $\mathcal{RES}$s can generate sustainable energy and are sourced from natural resources. The following are the primary categories of $\mathcal{RE}$ sources (Figure \ref{fig101}).\\

	\begin{figure}[htb!]
		\centering
		\includegraphics[width=12cm,height=4cm]{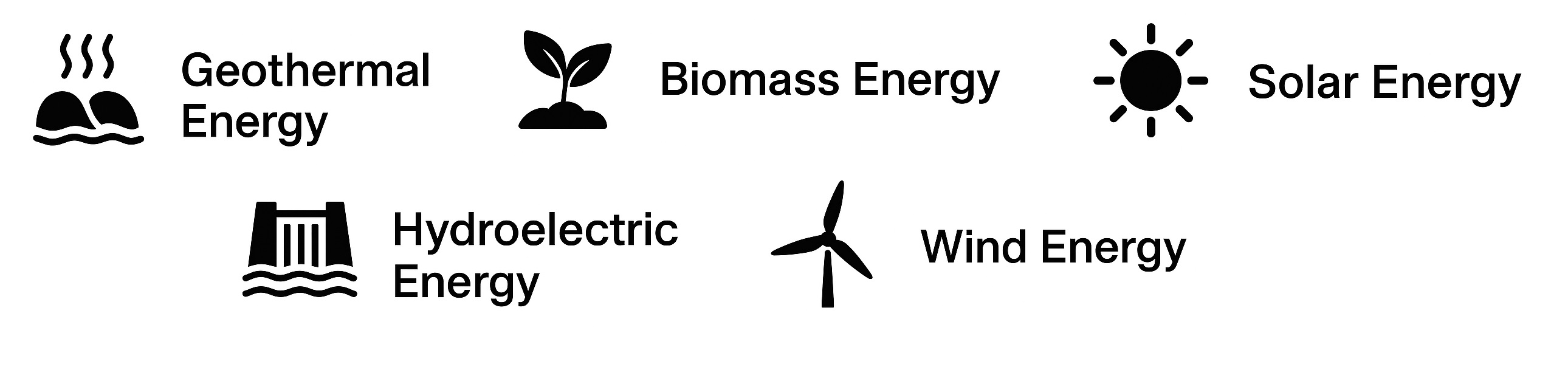}
		\caption{Renewable Energy Sources}\label{fig101}
	\end{figure}
	
	\textbf{Alternative S1: Geothermal Energy}\\
	
	Geothermal means ground heat. Water produced by natural processes—particularly precipitation—reaches the magma layer through fissures in the Earth's crust. The magma layer's heated water rises to the surface as steam and hot water. Turbines can turn the water and steam that reach the surface into energy. Geothermal energy is the thermal energy stored in the Earth's crust. This surface energy is transformed into electrical power by well-established power facilities. Additionally, it can be utilized in many patient-favored physical therapy facilities, tourist attractions, and central heating and cooling systems found in homes and workplaces. \\

		\textbf{Alternative S2: Bioenergy / Biomass Energy}\\

	This type of energy is an inexhaustible resource, readily available everywhere. It is considered a suitable and important energy source, particularly for rural areas, as it contributes to socio-economic development. Specially grown plants like corn and wheat, grasses, seaweed, marine algae, animal waste, fertilizer, industrial waste, and any organic waste (fruit and vegetable waste) left over from houses are biomass sources. The role of biomass in addressing the energy crisis is growing, despite the depletion of fossil fuels (coal, etc.) and the environmental harm they cause. \\

		\textbf{Alternative S3:  Solar Energy:}\\
		
	The sun is the solar system's energy source for all planets. It is an indispensable resource, especially for all living things on Earth. The practicality of their use, especially in the summer months, and the significant reductions they provide in electricity bills make solar panels a leading $\mathcal{RES}$ today. Solar panels, commonly found on the roofs of homes in cities and villages, can heat water, meet a home's hot water needs, or supply hot water to a home's heating system to meet heating needs. Solar energy technologies collect sunlight to generate heat or electricity. Solar energy is utilized in light, heat, and electricity. Solar energy systems convert the collected energy directly into electricity and can be installed on building roofs, appliances, and cars. Concentrated solar power plants use mirrors and lens systems to reflect solar radiation onto a relatively small area and generate electricity or heat.\\
	
	\textbf{Alternative S4: Hydroelectric Energy:} \\
	
	The basis of hydroelectric energy is to harness the energy of flowing water and convert it into electricity. Hydroelectric power plants are renewable and provide a clean energy source for nature. Because water flows at higher elevations are faster, these plants are more useful in these areas. Because hydroelectric power plants rely on the energy of flowing water, they are used to improve fishing, facilitate transportation, irrigate, and, most importantly, generate energy.\\

	\textbf{Alternative S5:	Wind Energy:}\\
	
	Solar energy is the source of wind energy. Wind is produced by the pressure differential that results from solar energy heating the land and the water at different rates. Wind turbines convert the kinetic energy of the wind into mechanical and electrical energy when positioned in windy areas. The present wind speed and duration determine how much energy the wind produces. Currently, 2\% of the world's electrical needs are met by wind. Compared to other power production methods, wind turbine technologies have a negligible environmental impact. \\

	Evaluating $\mathcal{RES}$s typically involves considering a range of factors and categories. This assessment covers many aspects, from energy production potential to environmental impacts. $\mathcal{RES}$s (solar, wind, hydroelectric, biomass, geothermal) can be analyzed from various perspectives, and each is evaluated under different criteria. Technical, environmental, social, and economic factors are all impacted differently by $\mathcal{RE}$ systems. The criteria details of these aspects can be introduced as follows:\\
	
	Economic aspects have three criteria: Investment cost(K1), 	O\&M(K2), Electric cost(K3); Technical aspects have three criteria: 	Efficiency(K4), Capacity factor(K5), and technical maturity(K6); Environmental aspects have two criteria: 	greenhouse gas (GHG) emission(K7), land use(K8); Social aspects have two criteria: 	Job creation(K9), social acceptance(K10).\\
	
	Investment cost: It covers the cost of the equipment and installation. Purchasing mechanical equipment, installing technology, building roads, connecting to the national web, hiring engineers, drilling, and other incidental construction activity are all examples of investment expenses \cite{wangCRIT}. The expenses and rewards of investments must be taken into account by the investors. The most popular economic criterion for assessing energy systems is investment cost.\\
	
	O\&M: It includes money spent on energy, goods, and services for the operation of the energy system, as well as personnel wages. It is divided into two sections: (1) operation costs, which include salaries for staff and the money spent on energy, goods, and services for the operation of the energy system \cite{wangCRIT}; and (2) maintenance costs. \\
	
	Electric Cost: It alludes to how much different energy generation systems cost. Land, construction, operation, fuel, equipment, and interest charges are all included in the total cost of the electrical power-producing process. This is an important financial sign as well.\\
	
	The ratio of energy input to energy production is known as efficiency. The most popular technical criterion for assessing energy systems is this one. The amount of sound energy extracted from an energy source is known as efficiency. The efficiency coefficient, or the ratio of output to input energy, is one of the most popular efficiency measures.  \\
	
	Capacity Factor: It displays the installed capacity divided by the annual total production. The ratio of electrical energy generated by a generating unit over a given period of time that might have been produced at continuous whole power operation during the same period is known as the power plant's capacity factor \cite{stein}. There are significant differences in the capacity factors of various power plant types. The weather impacts RE's capacity factor; for instance, hydropower plants' average electricity generating capacity is higher in the summer, which raises the capacity factor. While solar energy is influenced by sunlight and wind energy is influenced by wind speed, geothermal and biomass energy are operational nearly around the clock. \\

	Technical maturity: Technology relates to how dependable the adopted technology is and how widely it is used nationwide. This criterion has to deal with the accepted technology's level of dependability and national adoption. A qualitative 5-point rating system is used in this study to evaluate the technology, with five denoting very high maturity (i.e., commercially mature technology with a strong market position) and 1 denoting very low maturity (i.e., only laboratory testing). \\
	
	GHG emission: It refers to the greenhouse gas emissions created by the technology throughout its life cycle. One of the most popular metrics for assessing the stainability of $\mathcal{RES}$s is the quantity of greenhouse gas emissions from a particular $\mathcal{RE}$ system. Here, it is necessary to assess the life-cycle of greenhouse gas emissions from various renewable technologies. Emissions are expressed as g CO2eq/kWh, or equivalent emissions of CO2 per energy unit produced [58]. A life-cycle emissions assessment considers the energy production system's extraction, refining, processing, transportation, construction, operation and maintenance, and deconstruction phases.  \\
	
	Land Use: The land area needed for the technology is defined by its land usage. Because the energy system directly impacts the environment and landscape, the quantity of land needed for each plant is a crucial factor in the evaluation. Varying energy systems may occupy varying amounts of land, even when the products are identical. Thus, land usage needs to be taken into account.\\
	
	Job creation: The energy project is expected to generate potential job opportunities. An energy supply system might have a large workforce from construction to operation. Therefore, when developing RE, the inhabitants should be considered by evaluating the enhancement of their quality of life and creating jobs. \\

Social acceptance: It shows that the general public accepts $\mathcal{RE}$ technology. This criterion relates to the degree of public acceptance of RE, which is crucial in determining how RE technologies are implemented and reaching energy policy goals. It is crucial because public opinion and pressure groups can significantly impact how long an energy project takes to finish. It is important to remember that social approval cannot be measured directly. A qualitative scale, with one denoting strong resistance and five denoting strong support, is used in this study to quantify social acceptance.

\begin{table}[htb!]
	\centering
	\caption{Scale Values according to IVFF}\label{table01}
	\begin{tabular}{lllll}
		\hline
		Linguistic Terms & $\zeta_{L}$ & $\zeta_{U}$ & $\eta_{L}$ & $\eta_{U}$\\ \hline
		Certainly High Importance(CH) & 0.95 & 1 & 0 & 0\\
		Very High Importance(VH) & 0.8 & 0.9 & 0.1 & 0.2\\
		High Importance(H) & 0.7 & 0.8 & 0.2 & 0.3\\
		Slightly More Importance(SM) & 0.6 & 0.65 & 0.35 & 0.4\\
		Equally Importance(E) & 0.5 & 0.5 & 0.5 & 0.5\\
		Slightly Less Importance(SL) & 0.35 & 0.4 & 0.6 & 0.65\\
		Low Importance(L) & 0.2 & 0.3 & 0.7 & 0.8\\
		Very Low Importance(VL) & 0.1 & 0.2 & 0.8 & 0.9\\
		Certainly Low Importance(CL) & 0 & 0 & 0.95 & 1\\ \hline
	\end{tabular}
\end{table}


\subsection{Computations}

For this study, four $\mathcal{DMR}$s were interviewed to assess criteria and options associated with the RES selection dilemma. The first $\mathcal{DMR}$ has 29 years of experience and is an Electrical and Electronics Engineering professor. The second $\mathcal{DMR}$ is an engineer who has worked for the Ministry of Energy and Natural Resources of the Republic of Turkey for 20 years at the General Directorate of Energy Affairs. A worldwide bank employs the third $\mathcal{DMR}$, an electrical engineer with 16 years of experience, as an Energy Efficiency Specialist. The fourth $\mathcal{DMR}$ is an engineer who has worked for a private corporation in the energy industry for 11 years. Assume that the criteria weights are entirely unknown and that $\mathcal{DMR}$s' weight vector is $\lambda = (0.33, 0.28, 0.22, 0.17)$. It is expected that IVPFNs represent the assessment values of alternatives regarding criteria provided by the $\mathcal{DMR}$s, as shown in the IVFF group-$\mathcal{DCSM}$ provided in Table \ref{table101}.

\begin{table}[htb]
	\centering
	\caption{IVFF group decision matirx}\label{table101}
	\begin{tabular}{llllll|llllll}
		\hline
		& & U1  &  &  & & &  & U2 &  & &  \\ \hline
		Criteria & S1 & S2 & S3 & S4 & S5 & Criteria & S1 & S2 & S3 & A4 & S5 \\ \hline
		K1 & CH & SM & L & H & VH & K1 & VH & SM & SL & VH  & CH \\
		K2 & SM & H & L & CH & CH & K2 & E& H & VL & VH & CH\\
		K3 & CH & SM & SL & H & H &K3 & CH & E & L & H & SM\\
		K4 & SM & H & CH & SM & E &K4 & H& H & VH & SM & E\\
		K5 & L & SM & E & VH & CH &K5 & VL & E & E & CH & VH\\
		K6 & CH & VH & CH & H & SM &K6 & VH& VH & VH & H & H\\
		K7 & CH & VL & VL & SL & H &K7 & CH& L & VL & L & H\\
		K8 & SM & H & CH & H & SL &K8 & SM & VH & CH & H & E\\
		K9 & CH & CL & L & VL & SL &K9 & CH & VL & VL & CL & SL\\ 	
		K10 &  VH & VH & H & SM & H  &   K10 & VH& CH & VH & SM & H\\ 	\hline \hline
		&  & U3 &  &  & & &  & U4 &  & &  \\ \hline
		Criteria & S1 & S2 & S3 & S4 & S5 & Criteria & S1 & S2 & S3 & S4 & S5 \\ \hline
		K1 & CH  & H & SL & H & VH &   K1 & CH & SM & VL & H & SM\\
		K2 & L  & E & SM & H & SL &   K2 & E & H & SL & H & VH\\
		K3 &  VH  & SM & VL & SM & E &  K3 & CH & L & VL & E & SM\\
		K4 &  E  & SM & H & VH & E &  K4 & SM & H & VH & H & SM\\
		K5 &  CL  & SL & SL & H & H &  K5 & VL& SM & SL & VH & CH\\
		K6 &  H  & H & VH & SM & E &  K6 & CH & VH & VH & H & SM\\
		K7 &  VH  & VL & CL & SL & H &  K7 &CH & SL & L & SL & VH\\
		K8 &  SL  & H & VH & SM & L &  K8 & E & H & CH & E & VL\\
		K9 &  VH  & CL & CL & VL & SL &  K9 & CH& CL & VL & VL & E\\
		K10 &  H & VH & VH & E & SM &   K10 & VH& CH & SM & SL4 & E\\  		 \hline
	\end{tabular}
\end{table}

Equation \ref{optimal} is used to determine the criteria weights for each $\mathcal{DMR}$. 

\begin{eqnarray*}
	\omega^{1}&=&(0.214, 0.091, 0.067, 0.025, 0.125, 0.206, 0.14, 0.027, 0.045, 0.06)^{T}\\
	\omega^{2}&=&(0.307, 0.045, 0.033, 0.157, 0.052, 0.188, 0.096, 0.067, 0.023, 0.032)^{T}\\
	\omega^{3}&=&(0.153, 0.102, 0.041, 0.100, 0.029, 0.079, 0.108, 0.036, 0.012, 0.016)^{T}\\
	\omega^{4}&=&(0.225, 0.059, 0.107, 0.018, 0.091, 0.167, 0.211, 0.055, 0.042, 0.025)^{T}
\end{eqnarray*}

To determine the weights of the group's criterion, a linear programming model will be constructed (Model \ref{model4}).

\begin{eqnarray*}
	\min E(\omega^{*}) &=& 0.33 \times (\phi_{1}^{1} + \varphi_{1}^{1} + \cdots + \phi_{10}^{1} + \varphi_{10}^{1}) \\
	&+&  0.28 \times (\phi_{1}^{2} + \varphi_{1}^{2} + \cdots + \phi_{10}^{2} + \varphi_{10}^{2})\\
	&+& 0.22 \times (\phi_{1}^{3} + \varphi_{1}^{3} + \cdots + \phi_{10}^{3} + \varphi_{10}^{3})\\
	&+&  0.17 \times (\phi_{1}^{4} + \varphi_{1}^{4} + \cdots + \phi_{10}^{4} + \varphi_{10}^{4}),\\
\end{eqnarray*}
\begin{eqnarray}\label{model5}
\textit{s.t.} \quad \quad && 	0.000 - \omega_{1}^{*} - \phi_{1}^{1} + \varphi_{1}^{1}=0;\\ \nonumber
&&	0.000 - \omega_{2}^{*} - \phi_{2}^{1} + \varphi_{2}^{1}=0;\\ \nonumber
&&	0.000 - \omega_{3}^{*} - \phi_{3}^{1} + \varphi_{3}^{1}=0;\\ \nonumber
&&	\vdots \quad \vdots \quad \vdots \quad \vdots \\ \nonumber
&&	0.000 - \omega_{1}^{*} - \phi_{1}^{2} + \varphi_{1}^{2}=0;\\ \nonumber
&&	0.000 - \omega_{2}^{*} - \phi_{2}^{2} + \varphi_{2}^{2}=0;\\ \nonumber
&&		\vdots \quad \vdots \quad \vdots \quad \vdots \\ \nonumber
&&	0.000 - \omega_{1}^{*} - \phi_{1}^{3} + \varphi_{1}^{3}=0;\\ \nonumber
&&	0.000 - \omega_{2}^{*} - \phi_{2}^{3} + \varphi_{2}^{3}=0;\\ \nonumber
&&	\vdots \quad \vdots \quad \vdots \quad \vdots \\ \nonumber
&&	0.000 - \omega_{1}^{*} - \phi_{1}^{4} + \varphi_{1}^{4}=0;\\ \nonumber
&&	0.000 - \omega_{2}^{*} - \phi_{2}^{4} + \varphi_{2}^{4}=0;\\ \nonumber
&&	\vdots \quad \vdots \quad \vdots \quad \vdots \\ \nonumber
&&	0.000 - \omega_{10}^{*} - \phi_{10}^{4} + \varphi_{10}^{4}=0\\ \nonumber
&& \phi_{j}^{k} \geq 0,\\ \nonumber
&& \varphi_{j}^{k} \geq 0,\\ \nonumber
&& \phi_{j}^{k}\varphi_{j}^{k}=0, \quad j=1,2, \cdots, 10, \quad k=1,2,3,4\\ \nonumber
&&  \omega_{1}^{*} + \cdots + \omega_{10}^{*}=1\\ \nonumber
&&  \omega_{j}^{*} > 0 \quad j=1,2, \cdots, 10.
\end{eqnarray}

When model \ref{model5} is solved, then
 \begin{eqnarray*}
 	\omega^{*}=(0.221, 0.07, 0.14, 0.06, 0.074, 0.19, 0.14, 0.042, 0.03, 0.033)^{T}
 \end{eqnarray*}

In the meantime, we combine all of the individual $\mathcal{DCSM}$s into the collective $\mathcal{DCSM}$ using Equation \ref{IVFFWA}. The outcomes are indicated in Table \ref{table102}.\\

\begin{table}[htb!]
	\centering
\tiny
	\caption{Aggregated $\mathcal{DCSM}$}\label{table102}
	\begin{tabular}{llllll}
		\hline
		 & S1 & S2 & S3 & S4 & S5  \\ \hline
		K1 & ([0.52, 0.00], [0.00, 0.00])  &  ([0.67, 0.74], [0.23, 0.30])  &  ([0.34, 0.34], [0.85, 0.90]) &  ([0.63, 0.72], [0.33, 0.45])  &  ([0.52, 0.00], [0.00, 0.00]) \\
		K2 & ([0.43, 0.46], [0.66, 0.71])   &  ([0.70, 0.80], [0.18, 0.28]) & ([0.41, 0.40], [0.80, 0.84])   &  ([0.56, 0.00], [0.00, 0.00])  &  ([0.60, 0.00], [0.00, 0.00]) \\
		K3 &  ([0.54, 0.00], [0.00, 0.00]) & ([0.56, 0.60], [0.38, 0.42])   &   ([0.34, 0.34], [0.94, 0.95]) &  ([0.49, 0.55], [0.60, 0.66])  & ([0.73, 0.77], [0.20, 0.26])  \\
		K4 &  ([0.43, 0.46], [0.00, 0.00])& ([0.60, 0.67], [0.40, 0.50])  &  ([0.68, 0.00], [0.00, 0.00]) &    ([0.34, 0.37], [0.90, 0.91]) &   ([0.31, 0.31], [0.96, 0.97]) \\
		K5 & ([0.31, 0.31], [0.90, 0.95]) & ([0.50, 0.52], [0.60, 0.63]) &  ([0.34, 0.34], [0.94, 0.95]) &  ([0.68, 0.00], [0.00, 0.00])  &   ([0.54, 0.00], [0.00, 0.00])\\
		K6 & ([0.81, 0.00], [0.00, 0.00])  & ([0.68, 0.81], [0.27, 0.40]) &  ([0.66, 0.00], [0.00, 0.00])  & ([0.61 0.70], [0.37, 0.47]) & ([0.40, 0.46], [0.79, 0.83]) \\
		K7 & ([0.84, 0.00], [0.00, 0.00])  & ([0.34, 0.34], [0.82, 0.87])  & ([0.31, 0.31], [0.94, 0.95]) & ([0.34, 0.34], [0.67, 0.72])  &    ([0.40, 0.46], [0.79, 0.83]) \\
		K8 & ([0.34, 0.34], [0.91, 0.93])  &  ([0.37, 0.45], [0.83, 0.88]) & ([0.60, 0.00], [0.00, 0.00])  &   ([0.40, 0.46], [0.79, 0.83]) &   ([0.34, 0.34], [0.90, 0.92]) \\
		K9 &   ([0.73, 0.00], [0.00, 0.00]) &   ([0.21, 0.21], [0.96, 1.00]) & ([0.31, 0.31], [0.96, 0.97])  & ([0.31, 0.31], [0.96, 0.97])  & ([0.31, 0.31], [0.94, 0.95])  \\ 	
		K10 &  ([0.51, 0.63], [0.60, 0.70]) &  ([0.65, 0.00], [0.00, 0.00]) & ([0.31, 0.40], [0.88, 0.91]) &  ([0.34, 0.34], [0.94, 0.96]) &   ([0.63, 0.72], [0.33, 0.45]) \\ 	\hline
	\end{tabular}
\end{table}

In order to acquire the comprehensive preference values of options as displayed in Table \ref{table103}, we lastly use Equation \ref{IVFFWG} to aggregate the collective decision data in Table \ref{table102}. Equation \ref{Nscr} calculates the scores of each choice's comprehensive preference values. We can determine the ranking of each alternative based on these score values, as indicated in Table \ref{table103}.

\begin{table}[htb!]
	\centering
	\caption{The aggregated preference values, scores, and ranking of alternatives}\label{table103}
	\begin{tabular}{llll}
		\hline
	Alternative	& Aggregated Preference Values  & Scores & Ranking  \\ \hline
		S1 & ([0.54, 0.58], [0.00, 0.00])   & 0.59 &   5 \\
		S2 & ([0.93, 0.00], [0.00, 0.47]) & 0.68 &     4\\
		S3 & ([0.91, 0.92], [0.90, 0.00])  & 0.70 &   3  \\
		S4 & ([0.85, 0.87], [0.36, 0.00]) & 0.81 &   1 \\
		S5 & ([0.85, 0.88], [0.68, 0.00])  & 0.75 &  2  \\	\hline
	\end{tabular}
\end{table}

According to Table \ref{table103}, $S_{4} > S_{5} > S_{3} > S_{2} > S_{1}$, making $S_{4}$—biomass energy—the best option. The suggested approach can assist the $\mathcal{DMR}$ in choosing $\mathcal{RES}$s and successfully handle the $\mathcal{RES}$s selection problem. Furthermore, other MCGDM problems with partial weight information under IVFF contexts can also be solved using the suggested approach.

\section{Discussion}

The previous section ranked $\mathcal{RES}$s. This ranking resembles the actual energy combination of many countries today (hydro remains dominant, wind and solar are rapidly rising, and biomass and geothermal are more niche sources). This assessment demonstrates a logical ranking regarding current capacity, feasibility, and policy priorities.\\

To evaluate such a ranking, it is necessary to look at two dimensions: Managerial(technical) and strategic(political).\\

From a managerial perspective, the assessment can be as follows: Hydroelectric energy contributes to grid stability with its high capacity and ease of energy storage, and has a long lifespan. According to the results, hydroelectric energy's first-place ranking indicates its existing robust infrastructure and high production capacity. Disadvantages of hydroelectric energy include its environmental impact and the risk of drought. Advantages of wind energy include its low operating costs and near-zero carbon emissions. Its second-place ranking in the current study demonstrates its acceptance as a strategic investment area, particularly in high-potential regions.. Disadvantages of wind energy include intermittent generation, noise, and location dependency. Solar energy is easy to install, modular, and has rapidly decreasing costs. The third-place ranking indicates that, despite its rapid growth, solar energy has not yet been able to provide a base load. Disadvantages of solar energy include the problem of nighttime generation, namely intermittent generation, and the need for storage. Its integration with waste management and continuous production is a strength of biomass energy. Logistical challenges associated with its raw materials and the emission generation it generates, despite being carbon neutral, are disadvantages of biomass energy. Considering its operational challenges, its fourth-place ranking in the study makes sense. Geothermal energy provides continuous, uninterrupted energy and is low in carbon. Its limited widespread applicability explains the fifth finding in the study. Disadvantages of geothermal energy include its limited local potential, high investment costs, and environmental risks.\\

From a strategic perspective, the evaluation can be made as follows: This ranking may reflect the responding $\mathcal{DMR}$s' country conditions or the priorities of their respective institutions. Turning to domestic resources, such as hydroelectric and wind energy, is consistent with reducing fossil fuel imports. This demonstrates the prioritization of energy security. Since solar energy ranked third in these results, it can be considered a $\mathcal{RES}$ that should be supported regarding climate policies. Biomass and geothermal energy, on the other hand, may indicate that it should be limited to local projects.

	\subsection{Robustness Analysis:}

Something is solid, strong, and well-made when said to be robust. An analytical technique must yield the same result every time and everywhere it is applied, whether in the pharmaceutical, engineering, commercial, economic, or industrial sectors. The idea of method robustness was created in order to achieve this. The robustness of a technique indicates its dependability under normal operating conditions and quantifies its capacity to tolerate minor, deliberate alterations in method parameters.  \\

One factor at a time or an experimental design assesses robustness. Since only one factor is altered and the others remain constant in the one factor at a time approach, the impact of altering one factor can be observed. Although the data is considerably easier to analyze, this strategy can take much longer than the experimental design approach, where the elements are explored simultaneously. \\

	The COPRAS method can simultaneously evaluate both beneficial and detrimental criteria. The $\mathcal{DM}$ process is transparent, and what is calculated and how is clearly defined at each step. It has a simple and understandable structure, providing a clear ranking of alternatives, thus enabling $\mathcal{DMR}$s to make the right decision. Therefore, it is used in areas such as comparing $\mathcal{RES}$s and projects, evaluating environmental projects, product and process optimizations, and identifying different transportation systems.\\

The following algorithm steps of the IVFF-COPRAS method will be used for Robustness Analysis.\\

		IVFF-COPRAS Decision Steps:\\
		
Step 1: The construction of the individual IVFF-$\mathcal{DCSM}$.\\

Step 2: The IVFF-weighted averaging operator is used to create the IVFF-$\mathcal{DCSM}$.\\

Step 3: Equation \ref{BC} is used to acquire the IVFF maximizing indices for the benefit criterion, while Equation \ref{CC} is used to compute the IVFF minimizing indices for the cost criterion. 
		
		\begin{eqnarray}\label{BC}
			a_{i} &=& \left( \left[\sqrt[3]{1-\prod_{j=1}^{l}\left(1-\zeta_{ijL}^{3}\right)^{\omega_{j}}}, \sqrt[3]{1-\prod_{j=1}^{l}\left(1-\zeta_{ijU}^{3}\right)^{\omega_{j}}}\right], \left[\prod_{j=1}^{l}\eta_{ijL}^{\omega_{j}},  \prod_{j=1}^{l}\eta_{ijU}^{\omega_{j}}\right]\right) \\ \label{CC}
			b_{i} &=& \left( \left[\sqrt[3]{1-\prod_{j=l+1}^{n}\left(1-\zeta_{ijL}^{3}\right)^{\omega_{j}}}, \sqrt[3]{1-\prod_{j=l+1}^{n}\left(1-\zeta_{ijU}^{3}\right)^{\omega_{j}}}\right], \left[\prod_{j=l+1}^{n}\eta_{ijL}^{\omega_{j}},  \prod_{j=l+1}^{n}\eta_{ijU}^{\omega_{j}}\right]\right)
		\end{eqnarray}
		where $j=1,2, \cdots, l$ shows benefit criteria and $j=l+1, \cdots, n$ denotes cost criteria.\\
		
		Step 4: Equation \ref{RD} is used to calculate the relative degrees:

		\begin{eqnarray}\label{RD}
			\xi_{i} &=& SC(a_{i} )+\frac{\sum_{i=1}^{m}SC(b_{i})}{SC(b_{i})\sum_{i=1}^{m}\frac{1}{SC(b_{i})}}
		\end{eqnarray}
		where $i=1,2, \cdots, m$.\\
		
		Step 5: Equation \ref{UD} is used to assess the utility degree of options:
		
		\begin{eqnarray}\label{UD}
			U_{i} &=& \frac{\xi_{i}}{\xi_{max}} \times 100.
		\end{eqnarray}
		
		Step 6: The option with the highest utility degree is the best one. \\

	Commonly, COPRAS uses criteria weights and alternative normalization to calculate each alternative's benefit rating for beneficial criteria and loss rating for non-beneficial criteria, then generates the final ranking. However, in real life, robustness analysis is performed to determine how small changes in input data can alter the ranking. This tests the decision model's reliability and the results' sensitivity.\\

	Robustness analysis using COPRAS is performed as follows: First, calculations are performed using the COPRAS algorithm. Then, scenario changes are defined in the parameters. This way, the criterion weights are adjusted up or down by a certain percentage. Measurement errors or uncertainties are introduced. Alternatives are added and removed. Thus, COPRAS is recalculated based on the newly generated scenarios. COPRAS is repeated for each variation. The method is considered robust if two alternatives remain the same in most scenarios. The decision is considered fragile if even small changes significantly alter the ranking.\\

	We will analyze the study's robustness using IVFF-COPRAS \cite{gorcun}. The IVFF-COPRAS results were identical to those obtained using the $\mathcal{MD}$ method. In order to test the solution obtained using the IVFF-COPRAS algorithm, each alternative was first removed from the issue separately. The ranking of the available options according to cost and benefit criteria is shown in Table  \ref{tab:8}. 	Table \ref{tab:8} shows that the IVFF-COPRAS solution has no rank reversal problem.
	
	\begin{table}[htb!]
		\caption{Robustness Analysis}\label{tab:8}
		\vglue2mm
		\centering
		{
			\begin{tabular}{  c | c | c | c | c}
				\hline
				Alternatives & Ranking 1 & Ranking 2 & Ranking 3 & Ranking 4\\
				\hline
				$S_{1}$	& - & - & - &-\\
				$S_{2}$ & 4 & - & - &-\\
				$S_{3}$ & 3 & 2 & - &-\\
				$S_{4}$ & 1 & 1 & 1 &-\\
				$S_{5}$ & 2 & 3 & 2 &2\\
				\hline											
		\end{tabular}}
	\end{table}

\subsection{Comparative Analysis}

First, the results of two studies (\cite{Kabak}, \cite{Ren}) with RES selection will be presented for the comparative analysis. Then, the ordering will be performed using the $\mathcal{MD}$ method based on intuitionistic fuzzy and Pythagorean fuzzy sets (\cite{weiMDM}, \cite{Liangetal}).\\

	\begin{table}[htb!]
	\caption{Comparative Analysis}\label{CompAna}
	\vglue2mm
	\centering
	{
		\begin{tabular}{  ll}
			\hline
			Method & Ranking \\
			\hline
		PF-MDM \cite{Liangetal}	& $S_{4} > S_{1} > S_{2} > S_{5} > S_{3}$\\
		IF-MDM \cite{weiMDM} & $S_{4} > S_{2} > S_{5} > S_{1} > S_{3}$\\
		BOCR - AHP \cite{Kabak} & $S_{4} > S_{3} > S_{5} > S_{1} > S_{2}$ \\
		AHP-TOPSIS \cite{Ren} & $S_{4} > S_{5} > S_{2} > S_{3} > nuclear$\\
		Proposed Method & $S_{4} > S_{5} > S_{3} > S_{2} > S_{1}$\\
			\hline											
	\end{tabular}}
\end{table}

\subsection{Implications}
Choosing $\mathcal{RES}$s is not just a technical matter but has significant managerial and policy implications. These impacts can vary depending on which source (solar, wind, hydro, biomass, geothermal, etc.) is chosen in the $\mathcal{DM}$ process.\\

\textbf{Managerial implications: }This proposed method and the ranking of $\mathcal{RES}$s provide clear information for investment strategies and risk management. When choosing renewable resources, energy companies must consider capital intensity, payback period, and technological risks. For example, solar panels may offer shorter-term profitability, while geothermal investments have longer-term and higher initial costs. This ranking increases operational efficiency and simplifies technology selection. Managers must also determine infrastructure optimization, maintenance strategies, and resource selection. For example, wind turbines are maintenance-intensive, while solar panels require less maintenance. Diversifying the portfolio provides a competitive advantage. By investing in different renewable resources, companies can diversify their energy portfolio. This diversity provides a more robust position against market fluctuations, improves financial performance, and creates attractiveness for investors. Choosing the right resource increases investor confidence and facilitates raising funds within sustainability criteria (environmental, social, and governance - ESG). This brings the concept of corporate sustainability to life and positively impacts the image of countries/institutions. A country's or company's choice of renewable resources directly impacts its brand image and stakeholder trust. For example, a shift toward low-carbon resources like solar and wind strengthens perceptions of environmental responsibility.\\


\textbf{Policy implications:} This proposed method and ranking of $\mathcal{RES}$s are essential to countries' energy security and, consequently, their independence. Thus, resource selection can reduce a country's dependence on energy imports. For example, Turkey's solar and wind potential investment can reduce fossil fuel imports and the current account deficit. It also offers rich opportunities for regional development and employment policies. Policymakers can support regional employment and local development through renewable resource selection. For example, wind turbine factories or solar panel assembly facilities can stimulate the regional economy. This can be beneficial for countries to formulate climate policies in line with international policies and fulfill their international obligations. Fulfilling international climate commitments, such as the Paris Agreement and the Green Deal, determines resource selection. Choosing low-carbon resources contributes to greenhouse gas emission reduction targets. Properly using a country's resources can facilitate incentive mechanisms, making regulatory processes easier. This is because the state can implement incentives (feed-in tariffs, tax deductions, investment credits) to prioritize specific resources. This allows internal and external investors to influence their preferences directly. Considering these rankings will play a decisive role in infrastructure network planning. Policymakers will develop policies for energy storage, smart grids, and a balanced energy mix based on the source selection. Selecting $\mathcal{RE}$ systems within the framework of environmentally sound policies will facilitate social acceptance and ensure public support. As is well known, while hydroelectricity may face environmental and social objections in some regions, solar projects generally enjoy higher social acceptance.\\

Managerial and policy implications of $\mathcal{RES}$s ranked and analyzed for robustness using the $\mathcal{MCDM}$ technique are shown in Figure 1.

\begin{figure}
		\centering
	\includegraphics[width=13cm,height=7cm]{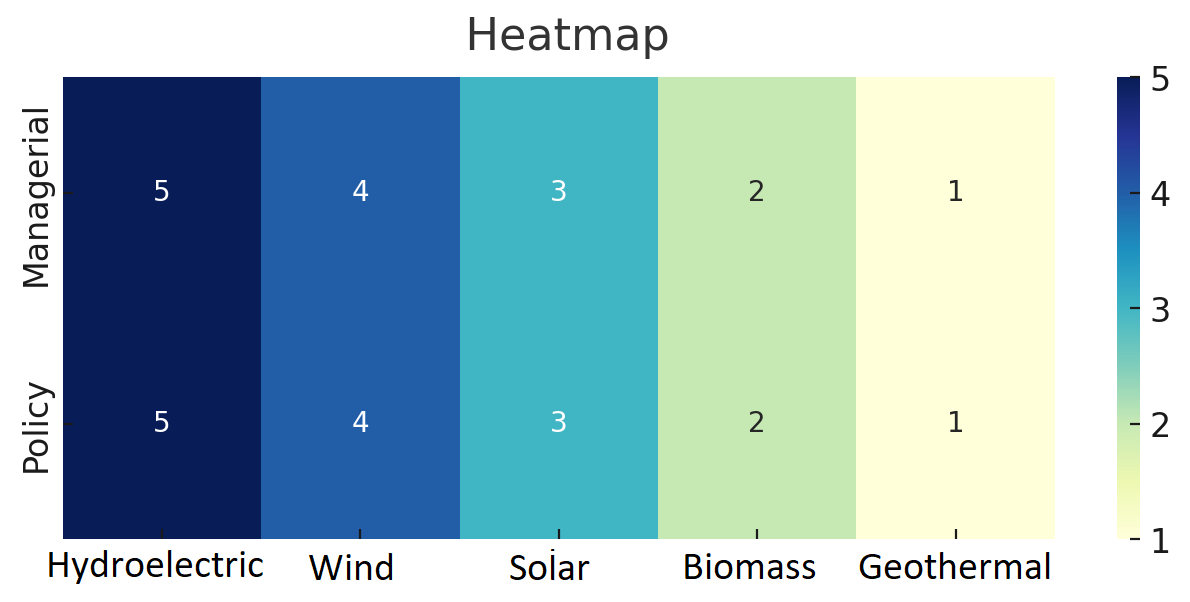}
	\caption{Heatmap}\label{fig00}
\end{figure}

\section{Conclusion}
Since IVFFS is a generalization of IVIFS and IVPFS, it can represent more ambiguous data, suggesting that it may be more suited to managing real-world $\mathcal{DM}$ applications. The environment's complexity and unpredictability mean that the weights of $\mathcal{DMR}$s and qualities are never fully understood or known. The $\mathcal{MD}$ approach is suggested to get over this restriction and tackle such $\mathcal{MCDM}$ issues. In this research, we explore the $\mathcal{MCDM}$ problems using IVFFSs by combining the IVFFS with the traditional $\mathcal{MD}$ method. We build two new $\mathcal{DM}$ models based on the conventional $\mathcal{MD}$ technique and the fundamental ideas of IVFFSs. Additionally, we have used the new models we suggested to tackle the $\mathcal{MCDM}$ problems using IVFF data. \\

In order to assess the choice of $\mathcal{RES}$s, an example is provided at the end, confirming the accuracy and scientific validity of the established model. The benefits of our proposed methods include expanding the breadth of assessment information and resolving $\mathcal{MCDM}$ difficulties with incomplete attribute weights. Though it is challenging to get relevant $\mathcal{DM}$ data because there are not enough $\mathcal{DMR}$s with an architectural background and fuzzy theory, the recently described models can produce more accurate $\mathcal{DM}$ results. \\

The rankings resulting from this study will provide significant guidance for administrative investment priorities, R\&D budgets, and infrastructure planning. It will also guide the determination of which resources are used to determine incentives, feed-in tariffs, local employment programs, and environmental regulations. Future studies could rank $\mathcal{RES}$s based on various factors. Different decision support systems, machine learning, or deep learning algorithms could be used for these rankings.








\end{document}